\documentclass{article}

\PassOptionsToPackage{numbers, compress}{natbib}
    \usepackage[final]{neurips_2024}

\usepackage[utf8]{inputenc} % allow utf-8 input
\usepackage[T1]{fontenc}    % use 8-bit T1 fonts
\usepackage{hyperref}       % hyperlinks
\usepackage{url}            % simple URL typesetting
\usepackage{booktabs}       % professional-quality tables
\usepackage{amsfonts}       % blackboard math symbols
\usepackage{nicefrac}       % compact symbols for 1/2, etc.
\usepackage{microtype}      % microtypography
\usepackage{xcolor}         % colors
\usepackage{soul}
\usepackage{graphicx}

\newcommand{\bb}[1]{{{#1}}} 

\title{Modeling cognitive processes of natural reading with transformer-based Language Models}

\author{%
  Bruno Bianchi$^{1,2,}$\thanks{https://liaa.dc.uba.ar/bruno-bianchi-en/} \\
  \texttt{bbianchi@dc.uba.ar} \\ 
\And
  Fermín Travi$^{1,2}$ \\
  \texttt{ftravi@dc.uba.ar} \\  
\And
  Juan E.~Kamienkowski$^{1,2,3}$ \\
  \texttt{jkamienk@dc.uba.ar} \\  % Affiliation 
\And 
    \\
  $^{1}$Universidad de Buenos Aires \\
  Facultad de Ciencias Exactas y Naturales\\
  Departamento de Computación \\
  Buenos Aires, Argentina. \\
\And
   \\
  $^{2}$Laboratorio de Inteligencia Artificial Aplicada \\
  CONICET-Universidad de Buenos Aires.\\
  Instituto de Ciencias de la Computación (ICC).\\
  Buenos Aires, Argentina.\\
\And
   \\
  $^{3}$Maestría en Explotación de Datos y \\
  Descubrimiento del Conocimiento\\
  Universidad de Buenos Aires \\
  Buenos Aires, Argentina. \\
}

\begin{document}

\maketitle

\begin{abstract}
Recent advances in Natural Language Processing (NLP) have led to the development of highly sophisticated language models for text generation. In parallel, neuroscience has increasingly employed these models to explore cognitive processes involved in language comprehension. Previous research has shown that models such as N-grams and LSTM networks can partially account for predictability effects in explaining eye movement behaviors, specifically Gaze Duration, during reading. In this study, we extend these findings by evaluating transformer-based models (GPT2, LLaMA-7B, and LLaMA2-7B) to further investigate this relationship. Our results indicate that these architectures outperform earlier models in explaining the variance in Gaze Durations recorded from Rioplantense Spanish readers. However, similar to previous studies, these models still fail to account for the entirety of the variance captured by human predictability. These findings suggest that, despite their advancements, state-of-the-art language models continue to predict language in ways that differ from human readers.

\end{abstract}

\section{Introduction}

Language is one of the defining characteristics of human beings. This unique capability allows us to communicate in complex ways to express thoughts \cite{hauser2002faculty}. Throughout history, various scientific fields—such as linguistics, psychology, and neuroscience—have studied language from different perspectives. In recent years, the field of Natural Language Processing (NLP) has gradually advanced toward algorithms and models capable of replicating human language with remarkable fidelity \cite{merityRegOpt2014,radford2018improving,radford2019language,brown2020language}. Although the primary aim of this field is not to understand language itself but rather to simulate it computationally, state-of-the-art models provide us with valuable tools that may contribute to understanding language in the brain. These tools complement ongoing efforts in cognitive neuroscience, which aims to elucidate the neural mechanisms underlying language. Ultimately, this creates a virtuous cycle where insights from both language and the brain’s mechanisms drive advancements in algorithm development \cite{toneva2019interpreting}.

In this context, researchers have examined how various algorithms from the field of NLP can estimate the human word-predictability. This variable reflects how likely it is that a human reader can anticipate a specific word before reading it, and it is widely known to have an impact on how words are processed. Predictability is typically measured using a task called the cloze-task, where participants are asked to complete sentences with a single word \cite{taylor1953cloze}. Predictability (cloze-Pred henceforth) is then calculated as the frequency with which participants choose the word that originally continues the sentence. 

Utilizing this variable to analyze various measurements of human cognition reveals its significant impact. For instance, studies measuring electroencephalography (EEG) signals during reading demonstrate that the brain's event-related potentials elicited by word processing (whether within sentences or in isolation) exhibit a marked correlation with the cloze-Pred of the presented words around 400 ms after word onset \cite{kutas1980reading}. This effect, known as the N400 (named for its negative polarity and 400 ms latency), has been extensively investigated over the past four decades \cite{kutas2011thirty}. Broadly speaking, the N400 reflects the cognitive effort required for processing the word read. Nonetheless, despite substantial efforts dedicated to this research over recent decades, its underlying causes are still widely discussed.

In a related but often independent field of study, eye movements are analyzed to gain deeper insights into cognitive processes during reading. In this domain, eye movements serve as an indirect yet reliable measure of the timing of cognitive processes. It is proposed, for instance, that the duration for which the eyes fixate on a word (which can vary) correlates with the time required by the brain to process that word  \cite{just1980theory}. Consequently, a substantial body of research in neurolinguistics focuses on reading studies that record fixation durations \cite{rayner1998eye, kliegl2004length,kliegl2006tracking, clifton2016eye}. These durations are subsequently employed in statistical models to analyze the variables influencing them. Such analyses, conducted in various forms over recent decades, have elucidated that certain properties of words, such as their length or lexical frequency, significantly impact fixation duration. From these correlations, it is hypothesized that length and lexical frequency influence word processing. For example, the observation that more frequent words are fixated for shorter durations suggests they are easier to process, as individuals likely possess a mental lexicon where more frequently encountered words are more readily accessible \cite{kliegl2006tracking}. 

Regarding cloze-Pred, it has been widely demonstrated that it negatively correlates with fixation time \cite{kliegl2006tracking,kamienkowski2018cumulative}. In other words, more predictable words are fixated for shorter durations. This, in turn, has led to hypotheses about how predictions are generated in the brain and how these predictions facilitate the anticipation and acceleration of processing future stimuli, a concept that could extend to cognition beyond the realm of reading. 

However, due to the costs associated with obtaining reliable measurements of cloze-Pred, studies often analyze the same texts repeatedly (for which measurements have already been made). Consequently, the development of a model capable of generating computational Predictabilities similar to human responses could represent a significant advancement in the study of language processing in the brain. Additionally, studies aimed at developing such a model would provide deeper insights into the underlying mechanisms of predictions generated by these models \cite{bianchi2020human,umfurer2021using,toneva2019interpreting}. 

In recent decades, numerous studies have attempted to model cloze-Pred using computational models, referred to as computational predictabilities (comp-Pred). However, these efforts have achieved only partial success to date \cite{ong2008conditional,hofmann2017benchmarking, bianchi2020human,hofmann2021language,algan2021prediction,umfurer2021using,shain2024large,lopes2024language}. Recently, there has been a growing interest in employing deep learning models for this purpose. Hofmann and colleagues \cite{hofmann2017benchmarking,hofmann2021language} trained N-grams, Topic Models (LDA), and Recurrent Neural Networks using datasets from Wikipedia texts and movie subtitles. Their analysis of variance in fixation times attributed to these probabilities led to the conclusion that computational models can better account for eye movements than cloze-Pred. Furthermore, \citet{shain2024large} explored the relationship between eye movement metrics and predictions from various statistical language models (Ngrams, GPT2, GPT3, amoung others), revealing a logarithmic connection. Adopting a different methodology, \citet{lopes2024language} utilized cloze-Pred and comp-Pred generated by GPT2 and LLaMA to simulate eye movement metrics using the OB1-reader model \cite{snell2018ob1}. They found that the simulated metrics produced with comp-Preds outperformed those generated with cloze-Pred in explaining human eye movements.

Another approach to analyze this was taken by \citet{bianchi2020human} and \citet{umfurer2021using}. In these studies they did not directly analyze the variance explained by each type of predictability (Cloze or computational), as this approach could overlook the possibility that different variables capture distinct portions of the variability in eye movement data. Notably, \citet{bianchi2020human} observed that, unlike cloze-Pred, all the computational predictability measures (models used: N-grams, LSA, word2vec, FastText, AWD-LSTM) captured a portion of the variance originally explained by word lexical frequency. This effect was slightly reduced when using an LSTM-based language model trained on Spanish Wikipedia and fine-tuned on texts from a similar domain to the evaluation materials (narrative texts). Additionally, their findings also show that the residuals of the used statistical models(linear mixed models) contain variance explained by cloze-Pred. That is, the variance accounted for by the analyzed comp-Pred is not the same than the cloze-Pred.

We are currently in an era where transformer-based architectures dominate the field of NLP \cite{vaswani2017attention,radford2018improving,devlin2018bert}. The present work aims to extend our previous efforts, first by utilizing transformer-based architectures, and second by refining the training corpus. Specifically, we aim to generate comp-Preds using a GPT2 model trained in Spanish, fine-tuned with two specialized corpora: one from the same literary domain as the evaluation texts, and the other from the same regional variant of Spanish as the participants (Rioplatense Spanish). Additionally, we are also testing predictions from more modern Large Language Models, like Llama and Llama2 in their 7B versions.

\section{Methods} 

\begin{figure*}[h!]
    \centering
    \includegraphics[width=.9\textwidth]{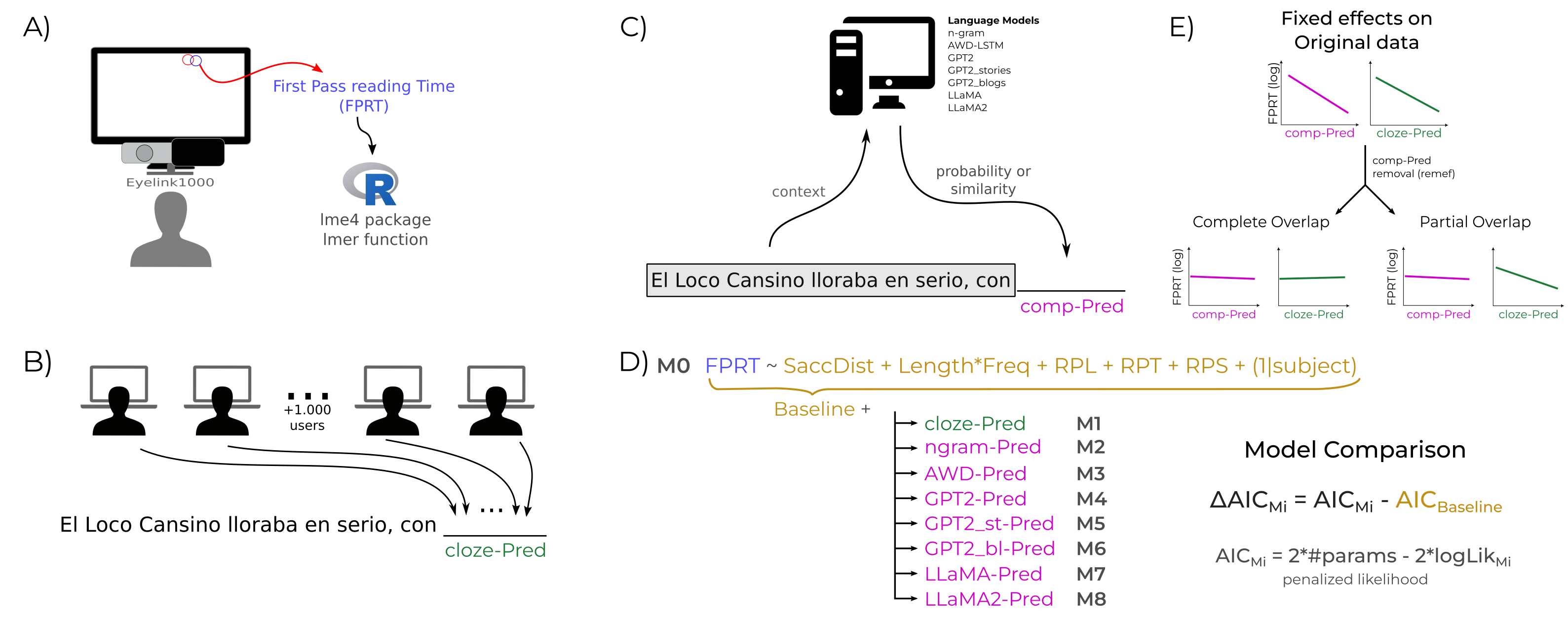}
    \caption{\bb{\textbf{Experimental designs: (A)}  Eye movements were recorded in native Spanish readers reading 8 narrative stories. The eye movement measure (FPRT) was used as dependent variable in statistical models in R. \textbf{(B)} cloze-Pred for texts used in (A) was estimated from online responses by +1,000 participants in a cloze-task experiment. \textbf{(C)} comp-Pred for the same texts was estimated with several Language Models. \textbf{(D)} Statistical models ran in the present study. cloze- and comp-Preds were individually added to a baseline model. Model comparison was performed using AIC difference. \textbf{(E)} Diagram of expected results for cloze and comp-Pred effects after the Remef analysis.}}
    \label{fig:method}
\end{figure*}

\subsection{Data: Texts, Cloze-Pred Estimation, and Eye Movements} 

In this study, we use the cloze-Pred and eye movement data previously published by Bianchi and colleagues \cite{bianchi2020human}. Eye movements (specifically, fixation durations) were recorded from 36 participants while reading 8 narrative texts. From these recordings, First-Pass Reading Time (FPRT: the sum of all fixations on a word during its first encounter) was calculated for each word read by each participant (Figure \ref{fig:method}A). This resulted in a total of $54,121$ data points, with an average of $1503\pm618$ per participant, $6765\pm3226$ per text, and $20\pm35$ fixations per word, covering $2,588$ unique words. FPRT will be used as the dependent variable in our statistical models. The corpus includes each of the 8 texts fully annotated with the following variables of interest: \textit{Saccadic Distance:} the number of characters traversed by the eye before the current fixation; \textit{Word Length:} the number of characters in the fixated word; \textit{Frequency:} lexical frequency of the word based on LexEsp; \textit{Rel pos:} relative positions in the line, text, or sentence; \textit{Length:Freq:} interaction between Word Length and Frequency; \textit{Pred:} Cloze and computational predictabilities (see below). For more details, see \cite{bianchi2020human}. 

\subsection{Human Predictability (cloze-Pred)} 
Data previously published by Bianchi and colleagues \citet{bianchi2020human} includes cloze predictability (cloze-Pred) values for all words in the dataset. Predictions for each word were gathered through an online cloze experiment involving approximately 1,000 participants (Figure \ref{fig:method}B). Each participant read at least one of the eight texts from the corpus, sequentially completing approximately one out of every thirty words. The cloze-Pred values were calculated based on an average of 13 responses per word, with a range of 8 to 37 responses. 

\subsection{Computational Predictability (comp-Pred)} 
\bb{To model cloze-Pred, we generated computational predictabilities (comp-Pred) using a Spanish version of GPT2 and two versions of LLaMA. These models were provided with the test texts to extract next-word probabilities corresponding to the completion of the original text (Figure \ref{fig:method}C).} 

In this study, we introduced the GPT2 language model architecture and compared it to previous models used by \citet{bianchi2020human} and \citet{umfurer2021using}. The model used was trained by the \textit{DeepEsp} consortium and is available through the \textit{HuggingFace} repository.\footnote{https://huggingface.co/DeepESP/gpt2-spanish} This model was trained on 11.5GB of Spanish text, comprising Wikipedia (3.5GB) and books (8GB) from various domains (narrative, short stories, theater, poetry, essays, and popular science). This pre-trained model will be used for two independent fine-tunings:

\begin{enumerate}
    \item For the first fine-tuning of this model, we used the corpus of narrative texts employed in previous studies \cite{bianchi2020human}. Due to licensing restrictions, this corpus is not publicly accessible. The corpus contains 2,082 stories (600MB), with a wide range of literary genres and authors from different nationalities.
    \item For the second fine-tuning texts sourced from Argentinian blogs  were used (28MB). This corpus is being prepared for public release. 
\end{enumerate} 

Both fine-tunings were performed on all parameters of the original model, using the code provided by HuggingFace for causal language model training.\footnote{https://github.com/huggingface/transformers/tree/main/examples/pytorch/language-modeling} Fine-tunings were performed using a Nvidia Titan RTX (VRAM 24Gb). The first fine-tuning took around 5 hours. The second one, less than an hour.

Finally, two models from the LLaMA \cite{touvron2023llama} family were used. Specifically, we employed the LLaMA and LLaMA2 models, both in their 7-billion-parameter versions. The pretrained versions from the HuggingFace repository were used.\footnote{https://huggingface.co/meta-llama} Contrary to GPT2, their pretraining corpus was multilingual.

\subsection{Linear Mixed Models} 
We employed Linear Mixed Models (LMMs, \textit{lme4} v.3.1-144, R v.3.6.3 \cite{bates2014lme4}) to analyze the log-transformed First Pass Reading Time (FPRT) variable. This variable is proposed to capture effects related to the early processing of words \cite{inhoff1984two}. Each of the fitted LMM returns a t-value for each co-variable, calculated by normalizing the estimated effect (slope) by its standard deviation (SD). Given the high number of data points at our disposal, we can consider infinite degrees of freedom. Thus, co-variables with $|t-value| > 2$ are considered to have a significant effect on the dependent variable. That is, under an assumed normal distribution, those estimated slopes that are more than 2 SD away from 0 have less than a 5\% probability of being 0 (see \cite{bianchi2020human} for more details). All the code necessary to run these statistical tests is openly available\footnote{https://github.com/brunobian/Modeling-cognitive-processes-LMs-Neurips2024}.

Our starting point is a baseline model (M0, Figure \ref{fig:method}D) that includes variables typically used in the study of eye movements (saccadic distance; word length; frequency; rel pos; length:freq) \cite{bianchi2020human,kliegl2004length}. This model serves as a foundation for understanding the effects of both cloze and computational predictability (see \cite{bianchi2020human,umfurer2021using} for a discussion on these effects). 
 
For comparing each resulting Linear Mixed Model (LMM) to the baseline model (M0), we use the Akaike Information Criterion (AIC). AIC evaluates the relative information loss of a model compared to other models, with a penalty for adding co-variables; a model that loses less information (i.e., has a lower AIC) is considered to be of higher quality. It is important to note that AIC is a relative metric, not an absolute one, meaning it can only be used to compare nested models fitted to the exact same dataset. Therefore, we report the difference in AIC between each model and the baseline model ($\Delta$AIC): the smaller this value, the better the model's performance.

Lastly, we reanalyzed all the models by adding cloze-Pred to its residuals. To achieve this, we extracted the residuals of each model by removing the estimated fixed effects. These residuals were then used in a new linear mixed model, with Cloze-Pred as the only fixed effect, keeping the same random effect structure. The aim of this analysis is to analyze the overlap between the variance explained by cloze-Pred and each comp-Pred. If a given comp-Pred can explain all of the variance of cloze-Pred (i.e. the computational model predicts like human readers), the effect of cloze-Pred on the residuals must be close to zero , (Figure \ref{fig:method}E).

\section{Results} 

Tables \ref{tabla_previos} and \ref{tabla_nuevos} presents the results (t-values) of the Linear Mixed Models (LMMs) conducted using the (log) First Pass Reading Time as the dependent variable against various combinations of co-variables. Models M0 through M3 (Table \ref{tabla_previos}) correspond to models analyzed in previous work by \citet{bianchi2020human} and \citet{umfurer2021using}. M0 is considered the Baseline Model, which includes all co-variables unrelated to Predictability. All of them have been proved to have a significant effect on the FPRT, as shown by the t-values presented in the corresponding column. 

\begin{table}[h!] \centering \caption{\textbf{Previous Linear Mixed Model Results:} Each column represents a different linear model. All models were applied to the same dataset. The results indicate the t-value obtained for each co-variable across the models. The row labeled ``Cloze-Remef'' refers to the t-value of the Cloze-Pred co-variable when used as the sole predictor in an LMM fitted on the residuals of the original model. All these models were previously presented by \citet{bianchi2020human} and \citet{umfurer2021using}. The row labeled ``$\Delta$AIC'' refers to the difference between the AIC of each model and the AIC of M0.} 
\label{tabla_previos} 
    \small{
\begin{tabular}{rccccccccc}
    \toprule

\textbf{Co-variable} & \textbf{M0} & \textbf{M1} & \textbf{M2} & \textbf{M3}\\
\textbf{}            & Base        & Cloze       & Ngram       & AWD epubs \\
\hline \vspace{-4pt} \\
Saccadic Dist.  & 44.44  & 44.63  & 43.92  & 44.31  \\
Length (inv)    & -18.15 & -18.56 & -19.10 & -18.59 \\
Frequency (log) & -10.83 & -10.60 & -1.93  & -5.77  \\
Rel pos line    & 4.14   & 3.96   & 4.33   & 4.12   \\
Rel pos text    & -3.93  & -3.28  & -3.87  & -4.68  \\
Rel pos sntc    & -5.36  & -4.85  & -5.76  & -4.97  \\
Len:Freq        & 16.98  & 17.17  & 15.68  & 16.91  \\[5pt] 
Pred (logit)    & --     & -16.23 & -21.02 & -18.23 \\[5pt] 
$\Delta$AIC     & 0      & -254   & -425   & -274  \\ [2pt]
\hline \vspace{-5pt}\\
Cloze-Remef     & -16.14 & 0.00   &  -9.47 & -7.37  \\

\bottomrule
\end{tabular}
}\normalsize
\end{table}

M1 incorporates cloze-Pred, representing the variable we aim to approximate with comp-Pred. As expected, this co-variable shows a significant negative effect, indicating that more predictable words are fixated for shorter durations. Notably, the inclusion of cloze-Pred has minimal to no impact on the effects of the other variables, suggesting that it explains a portion of the FPRT variance that was not captured by any other co-variable in M0. Additionally, this model has a better fit to the data than M0, according to the difference of their AICs ($\Delta\mbox{AIC}=-254$).

M2 represents the linear mixed model in which predictions from the best Ngram language model trained by \citet{bianchi2020human} are added to the M0. As discussed by \citet{umfurer2021using}, this simple model, based solely on counting the frequency of word chains of length N, generates one of the most human-like predictabilities to date. In the corresponding column of the table, it can be observed that the effect of this comp-Pred is greater than that of cloze-Pred, and is also negative. However, the addition of this co-variable caused significant changes in other effects, primarily in the Frequency effect. This suggests that the Ngram predictions account for variance that was previously attributed to Lexical Frequency. This is likely because the Ngram model, which counts word occurrences, is highly correlated with word frequency. Regarding the overall M2 performance, the AIC difference with M0 shows that M2 outperforms M1. That is, the co-variables combinations of M2 better explain the variance of the dependent variable.

Finally, M3 represents the best result obtained in \citet{umfurer2021using}, where an AWD-LSTM network \cite{merityRegOpt2014} was trained on Wikipedia and fine-tuned with a large corpus of narrative texts. Similar to the effect of Ngram comp-Pred, the AWD-LSTM comp-Pred effect is negative and larger than that of cloze-Pred. Additionally, it introduces significant, though smaller, changes to the other effects in the baseline model. In this case, the AIC analysis shows a slightly better fit to the data than M1, but worse than M2.

The ``Cloze-Remef'' row shows the t-value for cloze-Pred, obtained after fitting a new LMM on the residuals from each model, using only this co-variable. This approach allows us to analyze how much variance remaining in the residuals can be explained by the cloze-Pred. Trivially, in M0, the effect of cloze-Pred on the residuals is almost identical to its effect in M1. In contrast, in M1, there is no remaining variance in the residuals that can be explained by cloze-Pred. However, residuals of M2 and M3 contains some unexplained variance that can be accounted for by cloze-Pred, with the LSTM network’s comp-Pred leaving the least unexplained variance. It is noteworthy how the overall performance of each LMM on the data, measured with AIC, does not reflect the similarity of the corresponding comp-Pred with cloze-Pred.

Models M4 through M8 (Table \ref{tabla_nuevos}) correspond to novel models, where several versions of GPT2 and LLaMA Language Models are tested. M4 through M6 represent the statistical models corresponding to the predictabilities from the GPT2 variations. In M4, we introduced comp-Pred as computed with the pre-trained GPT2 model with no fine-tuning, downloaded directly from the HuggingFace repository. M5 and M6 introduce comp-Pred computed through the GPT2 model fine-tuned on short stories and Argentine blogs, respectively. Models M7 and M8 represent the statistical models corresponding to the predictabilities from LLaMA-7B and LLaMA2-7B models.

\begin{table}[h!] \centering \caption{\textbf{Novel Linear Mixed Model Results:} Each column represents a different linear model. All models were applied to the same dataset. The results indicate the t-value obtained for each co-variable across the models. The row labeled ``Cloze-Remef'' refers to the t-value of the Cloze-Pred co-variable when used as the sole predictor in an LMM fitted on the residuals of the original model. The row labeled "$\Delta$AIC" refers to the difference between the AIC of each model and the AIC of the M0.} 
\label{tabla_nuevos} 
    \small{
\begin{tabular}{rccccccccc}
\toprule
\textbf{Co-variable} & \textbf{M0} & \textbf{M4} & \textbf{M5} & \textbf{M6}& \textbf{M7}& \textbf{M8}\\
\textbf{} & Base & GPT2 & GPT2 stories & GPT2 blogs & Llama-7B & Llama2-7B\\
\hline \vspace{-4pt} \\
Saccadic Dist.  & 44.44  & 44.03  & 44.01  & 44.00  & 44.42  & 44.31\\
Length (inv)    & -18.15 & -20.04 & -19.90 & -19.90 & -18.25 & -19.93\\
Frequency (log) & -10.83 & -4.98  & -5.12  & -5.01  & -9.71  & -6.42\\
Rel pos line    & 4.14   & 4.45   & 4.56   & 4.61   & 4.22   & 4.25\\
Rel pos text    & -3.93  & -4.56  & -4.15  & -4.25  & -4.47  & -4.76 \\
Rel pos sntc    & -5.36  & -3.83  & -3.81  & -3.92  & -4.47  & -4.76\\
Len:Freq        & 16.98  & 15.59  & 15.58  & 15.61  & 17.20  & 16.43\\[5pt] 
Pred (logit)    & --     & -22.51 & -22.52 & -22.54 & -9.17  & -20.73\\[5pt] 
$\Delta$AIC & 0 & -482 & -482 & -482 & -81 & -407 \\[2pt] 
\hline \vspace{-5pt}\\
Cloze-Remef     & -16.14 & -6.18  & -6.39  & -6.58  & -13.19 & -6.12 \\
\bottomrule
\end{tabular}
}\normalsize
\end{table}

All three models based on GPT2 (M4, M5, and M6) show similar t-values for comp-Pred ($\mbox{t-val}_{Pred, M4-M6} \sim -22$) when compared to the Ngram-based model (M2) ($\mbox{t-val}_{Pred, M2}=-21,02$). However, when analyzing the impact of these comp-Preds on the other co-variables (compared to M0), we observe smaller variations than in M2. This is particularly evident in the effect of lexical Frequency ($\mbox{t-val}_{Freq, M0}=-10.83$). While in M2 lexical frequency loses significance ($\mbox{t-val}_{Freq, M2}=-1.93$), in models M4 to M6 the t-values remain above the significance threshold ($\mbox{t-val}_{Freq, M4-M6} \sim -5.00$). In this regard, the fine-tuned models seem to perform slightly better than the original model.

With respect to the residual variance analysis (Cloze-Remef), we observe that cloze-Pred is still able to capture enough variance to have a significant effect. Nevertheless, these remaining effects ($\mbox{t-val}_{Remef, M4-M6} \sim -6.40$) are smaller that the effects captured in the residuals of M2 ($\mbox{t-val}_{Remef, M2} = -9.47$) and M3 ($\mbox{t-val}_{Refemef, M3} = -7.37$).  

When comparing the comp-Pred derived from LLaMA models (M7 and M8), clear differences emerge between the two versions. The first version (LLaMA-7B) exhibits a smaller comp-Pred effect ($\mbox{t-val}{Pred, M7}=-9.17$), suggesting that its predictions are less aligned with cloze-Pred. Moreover, its influence on the Frequency effect is more subtle compared to the comp-Pred effects observed in previous models. Additionally, a significant amount of variance in this model's residuals can still be explained by cloze-Pred ($\mbox{t-val}{remef, M7}=-13.19$), indicating incomplete overlap between these predictions.

In contrast, the second version (LLaMA2) shows a marked improvement. The comp-Pred effect is more comparable to previous models ($\mbox{t-val}{Pred, M8}=-20.73$). However, this stronger effect comes with a notable reduction in the Frequency effect. Interestingly, LLaMA2, the most advanced model tested so far, leaves the least amount of unexplained variance for cloze-Pred ($\mbox{t-val}{remef, M8}=-6.12$), demonstrating its superior ability to account for the variance in predictability.

When it comes to overall model performance, with the exception of M7, where the first version of LLaMA produced suboptimal results, all models exhibited comparable AIC values. Notably, the model incorporating LLaMA2-7B predictabilities, which were the most similar to cloze-Pred, demonstrated slightly lower performance compared to the models using GPT2-based predictabilities.

\section{Discussion}

Recent advances in Natural Language Processing (NLP) models have opened new avenues for their application in Cognitive Neuroscience, helping to shed light on the underlying processes of language comprehension. Previous studies have made significant strides in understanding how these models, particularly causal language models, can estimate the likelihood of a human reader knowing a word before encountering it (i.e., cloze Predictability, or cloze-Pred) as a co-variable in statistical models analyzing eye movements \cite{bianchi2020human, umfurer2021using,hofmann2021language, hofmann2017benchmarking,lopes2024language}. In this study, we extended this research by evaluating the performance of transformer-based architectures in this context. \bb{Importantly, this research examined eye movement data from native Spanish readers analyzing Spanish texts. Despite Spanish being one of the most spoken languages globally, research on reading processes in Spanish remains underrepresented in the psycholinguistic literature.}

To this end, we analyzed the probabilities generated by the LLaMA-7B, LLaMA2-7B models, and three versions of the GPT2 model (a pre-trained version and two custom fine-tuned versions) as co-variables in the same linear mixed models used by \citet{bianchi2020human}. These models use First Pass Reading Time (FPRT, or Gaze Duration) as the dependent variable, along with a variety of co-variables corresponding to each word. The different computational predictability (comp-Pred) values generated by these models were added one at a time to the statistical models, which were then compared with the baseline model without any predictability (M0). Additionally, we analyzed the residuals of each model to assess the remaining variance that could be explained by cloze-Pred.

Results obtained in the present study shows that the GPT2 performance was independent of whether the model was used as-is from the HuggingFace repository or whether it was fine-tuned with domain-specific texts or the dialect variant of Spanish spoken by the readers. Two points about the fine-tuning are important to highlight. First, the original GPT2 model’s training data included narrative texts, so the fine-tuning may not have added much novel information to the model. Second, \bb{both fine-tunings were conducted with very small datasets (Rioplatense Spanish: 28MB, Narrative Stories: 600MB) compared to the original training corpus (11GB). Thus, it is possible that fine-tuning had little to no impact on the learned weights.}

Regarding the use of LLaMA family models, both the first and second versions in their 7-billion-parameter (7B) variants, the results show a significant difference between the two versions. Notably, the first version (analyzed in M7), released in 2023, performs worse than the GPT2 model, which was published four years earlier. This is evident not only in the smaller effect size of the corresponding comp-Pred but also in the residual variance that can be explained by cloze-Pred. However, when analyzing the second version of the LLaMA model (analyzed in M8), slightly better results are observed compared to GPT2, both in terms of residual variance and reduced interference with the Lexical Frequency effect. This leads to the hypothesis that more modern models are not only generating predictions that are somewhat closer to those of human readers, but they are also achieving this by becoming less dependent on the frequency of word occurrences in the lexicon.

The improvement in comp-Preds from transformer-based models, compared to previously used models (particularly AWD-LSTM), highlights the advantages of this architecture, which appears to capture richer linguistic information. Transformers benefit from processing entire input sequences simultaneously, allowing them to retain information from distant words without loss. Moreover, the increasing complexity of these models in recent years has further enhanced their predictive accuracy. This is largely driven by the significant growth in the number of internal parameters. For instance, while the AWD-LSTM model we used contains approximately $10^7$ parameters, both GPT2 and LLaMA models have $10^9$ parameters, representing a two-order-of-magnitude difference. \bb{This increase in the number of parameters, combined with recent optimizations to the transformer architecture, allows for enhanced extraction of intrinsic textual information, leading to improved language modeling and prediction capabilities. These improvements are evident both in the models' commercial applications as AI assistants and in our empirical findings. Our results demonstrate that transformer-based models generate predictions that rely less on lexical frequency compared to previous models.}

% As discussed throughout this paper, two distinctive aspects of our analysis are: 
% La diferencia principal entre nuestro analisis y lo que se hace en algunos trabajos previos es:

There are two main benefits to our analysis with respect to previous works \cite{hofmann2017benchmarking, hofmann2021language, shain2024large, lopes2024language}: (1) the interference generated by comp-Pred in the other co-variables of the models, and (2) the capacity of cloze-Pred to capture a portion of the residual variance. These analyses are essential for gaining a deeper understanding of computational predictions. Examining how these predictions account for variance in Fixation Duration in isolation from other effects could lead to overly optimistic conclusions. For instance, if the focus is solely on the observation that the effects of GPT2 or LLaMA2 are greater than that of cloze-Pred, or even that the AIC of the fitted LMMs are better, it might be concluded that predictions from state-of-the-art language models outperform those from human readers. However, our analysis reveals that this improvement comes at the expense of the Frequency effect (which has been extensively studied in the field of neurolinguistics) and leaving part of the cloze-Pred variance unexplained. \bb{In future work, it would be interesting to extend this analysis to other languages and datasets to determine whether the effects found in our Spanish corpus are replicated in the data analyzed in the related literature.}

In summary, this work contributes to the existing literature on predictions during reading, enhancing our understanding of both human predictability and computational predictabilities. Over the next few years, these two avenues must proceed hand in hand to foster synergy between the fields of cognitive neuroscience and artificial intelligence. This collaborative approach will enable us to combine resources and efforts to better understand the internal processes of both the human brain and state-of-the-art AI models, particularly in NLP. Furthermore, as advancements in AI continue to unfold, integrating insights from cognitive neuroscience could lead to the development of more sophisticated and human-like models. Ultimately, this intersection of disciplines holds the promise of unlocking new pathways for enhancing our comprehension of language processing, potentially paving the way for applications that not only improve AI performance but also offer deeper insights into human cognition.

\section{Limitations}

As previously mentioned, the corpora used for fine-tuning the GPT2 models are much smaller than the corpus used for pre-training. This may have resulted in the fine-tuning process not producing substantial changes in the pre-trained model's weights. However, it would be interesting to explore in greater depth whether it is possible to enrich these models with more information, both from the specific domain (narrative texts) and from the variant of Spanish spoken by the readers (Rioplatense Spanish). In the near future, we plan to expand the Rioplatense Spanish corpus to conduct a more thorough analysis of this type of fine-tuning. Additionally, we aim to further explore and diversify the analyses related to using these model outputs to improve our understanding of cognitive processes.

It is of great importance, both for the field of cognitive neuroscience and for artificial intelligence, that studies of this nature be conducted in as many languages and dialects as possible. Only in this way can we understand the general characteristics of our brain, while also ensuring equitable access to technological resources, such as today's generative NLP models.

\bibliographystyle{plainnat} 
\bibliography{biblio}

%%%%%%%%%%%%%%%%%%%%%%%%%%%%%%%%%%%%%%%%%%%%%%%%%%%%%%%%%%%%

% \appendix

% \section{Appendix / supplemental material}

% Optionally include supplemental material (complete proofs, additional experiments and plots) in appendix.
% All such materials \textbf{SHOULD be included in the main submission.}

%%%%%%%%%%%%%%%%%%%%%%%%%%%%%%%%%%%%%%%%%%%%%%%%%%%%%%%%%%%%

\newpage
\section*{NeurIPS Paper Checklist}

\begin{enumerate}

\item {\bf Claims}
    \item[] Question: Do the main claims made in the abstract and introduction accurately reflect the paper's contributions and scope?
    \item[] Answer: \answerYes{} % Replace by \answerYes{}, \answerNo{}, or \answerNA{}.
    \item[] Justification: The claims made in the abstract and introduction accurately reflect the paper's contributions and scope by clearly outlining the objectives of comparing state-of-the-art language models in predicting eye fixation durations during reading. Additionally, the integration of approaches from artificial intelligence and cognitive neuroscience is consistently emphasized throughout the paper, reinforcing the relevance of the findings to both fields.
    \item[] Guidelines:
    \begin{itemize}
        \item The answer NA means that the abstract and introduction do not include the claims made in the paper.
        \item The abstract and/or introduction should clearly state the claims made, including the contributions made in the paper and important assumptions and limitations. A No or NA answer to this question will not be perceived well by the reviewers. 
        \item The claims made should match theoretical and experimental results, and reflect how much the results can be expected to generalize to other settings. 
        \item It is fine to include aspirational goals as motivation as long as it is clear that these goals are not attained by the paper. 
    \end{itemize}

\item {\bf Limitations}
    \item[] Question: Does the paper discuss the limitations of the work performed by the authors?
    \item[] Answer: \answerYes{} % Replace by \answerYes{}, \answerNo{}, or \answerNA{}.
    \item[] Justification: The paper includes a dedicated Limitations section that addresses the constraints of the authors' work. For instance, it highlights the preference for using larger training datasets for fine-tuning to enhance model performance.
    \item[] Guidelines:
    \begin{itemize}
        \item The answer NA means that the paper has no limitation while the answer No means that the paper has limitations, but those are not discussed in the paper. 
        \item The authors are encouraged to create a separate "Limitations" section in their paper.
        \item The paper should point out any strong assumptions and how robust the results are to violations of these assumptions (e.g., independence assumptions, noiseless settings, model well-specification, asymptotic approximations only holding locally). The authors should reflect on how these assumptions might be violated in practice and what the implications would be.
        \item The authors should reflect on the scope of the claims made, e.g., if the approach was only tested on a few datasets or with a few runs. In general, empirical results often depend on implicit assumptions, which should be articulated.
        \item The authors should reflect on the factors that influence the performance of the approach. For example, a facial recognition algorithm may perform poorly when image resolution is low or images are taken in low lighting. Or a speech-to-text system might not be used reliably to provide closed captions for online lectures because it fails to handle technical jargon.
        \item The authors should discuss the computational efficiency of the proposed algorithms and how they scale with dataset size.
        \item If applicable, the authors should discuss possible limitations of their approach to address problems of privacy and fairness.
        \item While the authors might fear that complete honesty about limitations might be used by reviewers as grounds for rejection, a worse outcome might be that reviewers discover limitations that aren't acknowledged in the paper. The authors should use their best judgment and recognize that individual actions in favor of transparency play an important role in developing norms that preserve the integrity of the community. Reviewers will be specifically instructed to not penalize honesty concerning limitations.
    \end{itemize}

\item {\bf Theory Assumptions and Proofs}
    \item[] Question: For each theoretical result, does the paper provide the full set of assumptions and a complete (and correct) proof?
    \item[] Answer: \answerNA{} % Replace by \answerYes{}, \answerNo{}, or \answerNA{}.
    \item[] Justification: The paper does not include theoretical results.
    \item[] Guidelines:
    \begin{itemize}
        \item The answer NA means that the paper does not include theoretical results. 
        \item All the theorems, formulas, and proofs in the paper should be numbered and cross-referenced.
        \item All assumptions should be clearly stated or referenced in the statement of any theorems.
        \item The proofs can either appear in the main paper or the supplemental material, but if they appear in the supplemental material, the authors are encouraged to provide a short proof sketch to provide intuition. 
        \item Inversely, any informal proof provided in the core of the paper should be complemented by formal proofs provided in appendix or supplemental material.
        \item Theorems and Lemmas that the proof relies upon should be properly referenced. 
    \end{itemize}

    \item {\bf Experimental Result Reproducibility}
    \item[] Question: Does the paper fully disclose all the information needed to reproduce the main experimental results of the paper to the extent that it affects the main claims and/or conclusions of the paper (regardless of whether the code and data are provided or not)?
    \item[] Answer: \answerYes{} % Replace by \answerYes{}, \answerNo{}, or \answerNA{}.
    \item[] Justification: The paper provides sufficient information for reproducing the main experimental results that support its claims and conclusions. The eye movement data and cloze-Pred values are publicly available from the original authors, while the GPT2 and LLaMA models used are open access. Although the fine-tuning corpora are not published, the fine-tuned models did not significantly alter the results. Furthermore, some of this data will be published shortly. 
    \item[] Guidelines:
    \begin{itemize}
        \item The answer NA means that the paper does not include experiments.
        \item If the paper includes experiments, a No answer to this question will not be perceived well by the reviewers: Making the paper reproducible is important, regardless of whether the code and data are provided or not.
        \item If the contribution is a dataset and/or model, the authors should describe the steps taken to make their results reproducible or verifiable. 
        \item Depending on the contribution, reproducibility can be accomplished in various ways. For example, if the contribution is a novel architecture, describing the architecture fully might suffice, or if the contribution is a specific model and empirical evaluation, it may be necessary to either make it possible for others to replicate the model with the same dataset, or provide access to the model. In general. releasing code and data is often one good way to accomplish this, but reproducibility can also be provided via detailed instructions for how to replicate the results, access to a hosted model (e.g., in the case of a large language model), releasing of a model checkpoint, or other means that are appropriate to the research performed.
        \item While NeurIPS does not require releasing code, the conference does require all submissions to provide some reasonable avenue for reproducibility, which may depend on the nature of the contribution. For example
        \begin{enumerate}
            \item If the contribution is primarily a new algorithm, the paper should make it clear how to reproduce that algorithm.
            \item If the contribution is primarily a new model architecture, the paper should describe the architecture clearly and fully.
            \item If the contribution is a new model (e.g., a large language model), then there should either be a way to access this model for reproducing the results or a way to reproduce the model (e.g., with an open-source dataset or instructions for how to construct the dataset).
            \item We recognize that reproducibility may be tricky in some cases, in which case authors are welcome to describe the particular way they provide for reproducibility. In the case of closed-source models, it may be that access to the model is limited in some way (e.g., to registered users), but it should be possible for other researchers to have some path to reproducing or verifying the results.
        \end{enumerate}
    \end{itemize}

\item {\bf Open access to data and code}
    \item[] Question: Does the paper provide open access to the data and code, with sufficient instructions to faithfully reproduce the main experimental results, as described in supplemental material?
    \item[] Answer: \answerYes{} % Replace by \answerYes{}, \answerNo{}, or \answerNA{}.
    \item[] Justification: All the data used in this study comes from openly available datasets that have been published for public use, ensuring proper credit to the original creators and compliance with their licensing terms.
    \item[] Guidelines:
    \begin{itemize}
        \item The answer NA means that paper does not include experiments requiring code.
        \item Please see the NeurIPS code and data submission guidelines (\url{https://nips.cc/public/guides/CodeSubmissionPolicy}) for more details.
        \item While we encourage the release of code and data, we understand that this might not be possible, so “No” is an acceptable answer. Papers cannot be rejected simply for not including code, unless this is central to the contribution (e.g., for a new open-source benchmark).
        \item The instructions should contain the exact command and environment needed to run to reproduce the results. See the NeurIPS code and data submission guidelines (\url{https://nips.cc/public/guides/CodeSubmissionPolicy}) for more details.
        \item The authors should provide instructions on data access and preparation, including how to access the raw data, preprocessed data, intermediate data, and generated data, etc.
        \item The authors should provide scripts to reproduce all experimental results for the new proposed method and baselines. If only a subset of experiments are reproducible, they should state which ones are omitted from the script and why.
        \item At submission time, to preserve anonymity, the authors should release anonymized versions (if applicable).
        \item Providing as much information as possible in supplemental material (appended to the paper) is recommended, but including URLs to data and code is permitted.
    \end{itemize}

\item {\bf Experimental Setting/Details}
    \item[] Question: Does the paper specify all the training and test details (e.g., data splits, hyperparameters, how they were chosen, type of optimizer, etc.) necessary to understand the results?
    \item[] Answer: \answerYes{} % Replace by \answerYes{}, \answerNo{}, or \answerNA{}.
    \item[] Justification:  The Methods section provides all the necessary information regarding the fine-tuning of GPT2. The LLaMA versions used in this study were not fine-tuned.
    \item[] Guidelines:
    \begin{itemize}
        \item The answer NA means that the paper does not include experiments.
        \item The experimental setting should be presented in the core of the paper to a level of detail that is necessary to appreciate the results and make sense of them.
        \item The full details can be provided either with the code, in appendix, or as supplemental material.
    \end{itemize}

\item {\bf Experiment Statistical Significance}
    \item[] Question: Does the paper report error bars suitably and correctly defined or other appropriate information about the statistical significance of the experiments?
    \item[] Answer: \answerYes{} % Replace by \answerYes{}, \answerNo{}, or \answerNA{}.
    \item[] Justification:  Yes, the Methods section includes comprehensive information about the statistical analysis, including a clear explanation of t-value interpretation and references for further discussions on the topic. This ensures that statistical significance is reported appropriately.
    \item[] Guidelines:
    \begin{itemize}
        \item The answer NA means that the paper does not include experiments.
        \item The authors should answer "Yes" if the results are accompanied by error bars, confidence intervals, or statistical significance tests, at least for the experiments that support the main claims of the paper.
        \item The factors of variability that the error bars are capturing should be clearly stated (for example, train/test split, initialization, random drawing of some parameter, or overall run with given experimental conditions).
        \item The method for calculating the error bars should be explained (closed form formula, call to a library function, bootstrap, etc.)
        \item The assumptions made should be given (e.g., Normally distributed errors).
        \item It should be clear whether the error bar is the standard deviation or the standard error of the mean.
        \item It is OK to report 1-sigma error bars, but one should state it. The authors should preferably report a 2-sigma error bar than state that they have a 96\% CI, if the hypothesis of Normality of errors is not verified.
        \item For asymmetric distributions, the authors should be careful not to show in tables or figures symmetric error bars that would yield results that are out of range (e.g. negative error rates).
        \item If error bars are reported in tables or plots, The authors should explain in the text how they were calculated and reference the corresponding figures or tables in the text.
    \end{itemize}

\item {\bf Experiments Compute Resources}
    \item[] Question: For each experiment, does the paper provide sufficient information on the computer resources (type of compute workers, memory, time of execution) needed to reproduce the experiments?
    \item[] Answer: \answerYes{} % Replace by \answerYes{}, \answerNo{}, or \answerNA{}.
    \item[] Justification: The Methods section includes detailed information about the computer resources utilized for fine-tuning the GPT2 model, ensuring that readers have sufficient details to reproduce the experiments.
    \item[] Guidelines:
    \begin{itemize}
        \item The answer NA means that the paper does not include experiments.
        \item The paper should indicate the type of compute workers CPU or GPU, internal cluster, or cloud provider, including relevant memory and storage.
        \item The paper should provide the amount of compute required for each of the individual experimental runs as well as estimate the total compute. 
        \item The paper should disclose whether the full research project required more compute than the experiments reported in the paper (e.g., preliminary or failed experiments that didn't make it into the paper). 
    \end{itemize}
    
\item {\bf Code Of Ethics}
    \item[] Question: Does the research conducted in the paper conform, in every respect, with the NeurIPS Code of Ethics \url{https://neurips.cc/public/EthicsGuidelines}?
    \item[] Answer: \answerYes{} % Replace by \answerYes{}, \answerNo{}, or \answerNA{}.
    \item[] Justification: We have reviewed the NeurIPS Code of Ethics and conducted our research in accordance with all its guidelines.
    \item[] Guidelines:
    \begin{itemize}
        \item The answer NA means that the authors have not reviewed the NeurIPS Code of Ethics.
        \item If the authors answer No, they should explain the special circumstances that require a deviation from the Code of Ethics.
        \item The authors should make sure to preserve anonymity (e.g., if there is a special consideration due to laws or regulations in their jurisdiction).
    \end{itemize}

\item {\bf Broader Impacts}
    \item[] Question: Does the paper discuss both potential positive societal impacts and negative societal impacts of the work performed?
    \item[] Answer: \answerYes{} % Replace by \answerYes{}, \answerNo{}, or \answerNA{}.
    \item[] Justification: This paper discusses the need to conduct such analyses in underrepresented languages, at least in the literature, such as Spanish. We believe that this could enhance our understanding of our culture and lead to the development of better NLP models.
    \item[] Guidelines:
    \begin{itemize}
        \item The answer NA means that there is no societal impact of the work performed.
        \item If the authors answer NA or No, they should explain why their work has no societal impact or why the paper does not address societal impact.
        \item Examples of negative societal impacts include potential malicious or unintended uses (e.g., disinformation, generating fake profiles, surveillance), fairness considerations (e.g., deployment of technologies that could make decisions that unfairly impact specific groups), privacy considerations, and security considerations.
        \item The conference expects that many papers will be foundational research and not tied to particular applications, let alone deployments. However, if there is a direct path to any negative applications, the authors should point it out. For example, it is legitimate to point out that an improvement in the quality of generative models could be used to generate deepfakes for disinformation. On the other hand, it is not needed to point out that a generic algorithm for optimizing neural networks could enable people to train models that generate Deepfakes faster.
        \item The authors should consider possible harms that could arise when the technology is being used as intended and functioning correctly, harms that could arise when the technology is being used as intended but gives incorrect results, and harms following from (intentional or unintentional) misuse of the technology.
        \item If there are negative societal impacts, the authors could also discuss possible mitigation strategies (e.g., gated release of models, providing defenses in addition to attacks, mechanisms for monitoring misuse, mechanisms to monitor how a system learns from feedback over time, improving the efficiency and accessibility of ML).
    \end{itemize}
    
\item {\bf Safeguards}
    \item[] Question: Does the paper describe safeguards that have been put in place for responsible release of data or models that have a high risk for misuse (e.g., pretrained language models, image generators, or scraped datasets)?
    \item[] Answer: \answerNA{} % Replace by \answerYes{}, \answerNo{}, or \answerNA{}.
    \item[] Justification: Most of the data used in this work has already been published. The data we haven't released due to rights restrictions consists of widely disseminated texts that do not contain sensitive material that could be misused.
    \item[] Guidelines:
    \begin{itemize}
        \item The answer NA means that the paper poses no such risks.
        \item Released models that have a high risk for misuse or dual-use should be released with necessary safeguards to allow for controlled use of the model, for example by requiring that users adhere to usage guidelines or restrictions to access the model or implementing safety filters. 
        \item Datasets that have been scraped from the Internet could pose safety risks. The authors should describe how they avoided releasing unsafe images.
        \item We recognize that providing effective safeguards is challenging, and many papers do not require this, but we encourage authors to take this into account and make a best faith effort.
    \end{itemize}

\item {\bf Licenses for existing assets}
    \item[] Question: Are the creators or original owners of assets (e.g., code, data, models), used in the paper, properly credited and are the license and terms of use explicitly mentioned and properly respected?
    \item[] Answer: \answerYes{} % Replace by \answerYes{}, \answerNo{}, or \answerNA{}.
    \item[] Justification: All the data used in this study comes from openly available datasets that have been published for public use, ensuring proper credit to the original creators and compliance with their licensing terms.
    \item[] Guidelines:
    \begin{itemize}
        \item The answer NA means that the paper does not use existing assets.
        \item The authors should cite the original paper that produced the code package or dataset.
        \item The authors should state which version of the asset is used and, if possible, include a URL.
        \item The name of the license (e.g., CC-BY 4.0) should be included for each asset.
        \item For scraped data from a particular source (e.g., website), the copyright and terms of service of that source should be provided.
        \item If assets are released, the license, copyright information, and terms of use in the package should be provided. For popular datasets, \url{paperswithcode.com/datasets} has curated licenses for some datasets. Their licensing guide can help determine the license of a dataset.
        \item For existing datasets that are re-packaged, both the original license and the license of the derived asset (if it has changed) should be provided.
        \item If this information is not available online, the authors are encouraged to reach out to the asset's creators.
    \end{itemize}

\item {\bf New Assets}
    \item[] Question: Are new assets introduced in the paper well documented and is the documentation provided alongside the assets?
    \item[] Answer: \answerYes{} % Replace by \answerYes{}, \answerNo{}, or \answerNA{}.
    \item[] Justification: All the code utilized in this study is openly published on GitHub, ensuring thorough documentation is available alongside the assets.
    \item[] Guidelines:
    \begin{itemize}
        \item The answer NA means that the paper does not release new assets.
        \item Researchers should communicate the details of the dataset/code/model as part of their submissions via structured templates. This includes details about training, license, limitations, etc. 
        \item The paper should discuss whether and how consent was obtained from people whose asset is used.
        \item At submission time, remember to anonymize your assets (if applicable). You can either create an anonymized URL or include an anonymized zip file.
    \end{itemize}

\item {\bf Crowdsourcing and Research with Human Subjects}
    \item[] Question: For crowdsourcing experiments and research with human subjects, does the paper include the full text of instructions given to participants and screenshots, if applicable, as well as details about compensation (if any)? 
    \item[] Answer: \answerNA{} % Replace by \answerYes{}, \answerNo{}, or \answerNA{}.
    \item[] Justification: The data used from these sources were not obtained specifically for this study but rather sourced from publicly available datasets.
    \item[] Guidelines:
    \begin{itemize}
        \item The answer NA means that the paper does not involve crowdsourcing nor research with human subjects.
        \item Including this information in the supplemental material is fine, but if the main contribution of the paper involves human subjects, then as much detail as possible should be included in the main paper. 
        \item According to the NeurIPS Code of Ethics, workers involved in data collection, curation, or other labor should be paid at least the minimum wage in the country of the data collector. 
    \end{itemize}

\item {\bf Institutional Review Board (IRB) Approvals or Equivalent for Research with Human Subjects}
    \item[] Question: Does the paper describe potential risks incurred by study participants, whether such risks were disclosed to the subjects, and whether Institutional Review Board (IRB) approvals (or an equivalent approval/review based on the requirements of your country or institution) were obtained?
    \item[] Answer: \answerNA{} % Replace by \answerYes{}, \answerNo{}, or \answerNA{}.
    \item[] Justification: The data used from these sources were not obtained specifically for this study but rather sourced from publicly available datasets.
    \item[] Guidelines:
    \begin{itemize}
        \item The answer NA means that the paper does not involve crowdsourcing nor research with human subjects.
        \item Depending on the country in which research is conducted, IRB approval (or equivalent) may be required for any human subjects research. If you obtained IRB approval, you should clearly state this in the paper. 
        \item We recognize that the procedures for this may vary significantly between institutions and locations, and we expect authors to adhere to the NeurIPS Code of Ethics and the guidelines for their institution. 
        \item For initial submissions, do not include any information that would break anonymity (if applicable), such as the institution conducting the review.
    \end{itemize}

\end{enumerate}

\end{document}